\pdfoutput=1
\documentclass[11pt,a4paper]{article}
\usepackage[hyperref]{acl2019}
\usepackage{times}
\usepackage{latexsym}

\usepackage{booktabs}
\usepackage{multirow}
\usepackage{graphicx}
\usepackage{array}
\usepackage{tabularx}
\usepackage{url}
\usepackage{amsmath}
\usepackage{pifont}
\usepackage{tikz,pgfplots}
\usepackage[utf8]{inputenc}
\usepgfplotslibrary{groupplots}
\pgfplotsset{compat=1.14}

\usetikzlibrary{shapes,arrows, positioning}

\newcolumntype{N}{>{\centering\arraybackslash}X}
\newcolumntype{P}[1]{>{\centering\arraybackslash}p{#1}}
\newcolumntype{L}[1]{>{\raggedright\arraybackslash}p{#1}}

\definecolor{g-red}{HTML}{DB4437}
\definecolor{g-blue}{HTML}{4285F4}
\definecolor{g-green}{HTML}{0F9D58}
\definecolor{g-yellow}{HTML}{F4B400}
\definecolor{g-orange}{HTML}{FF9800}
\definecolor{g-grey}{HTML}{9E9E9E}

\usepackage[colorinlistoftodos,prependcaption,textsize=tiny]{todonotes}
\newcounter{kt}

\newcounter{mw}

\newcounter{kl}

\newcommand\yesmark{\ding{51}}

\tikzstyle{bert_q} = [draw, fill=g-green!20, rectangle, anchor=south, align=center, text width=12em, minimum height=3em]
\tikzstyle{bert_b} = [draw, fill=g-blue!20, rectangle, anchor=south, align=center, text width=12em, minimum height=3em]
\tikzstyle{bert_r} = [draw, fill=g-red!20, rectangle, anchor=south, align=center, text width=12em, minimum height=3em]
\tikzstyle{input} = [draw, align=center, text width=12em, anchor=north, node distance=0cm, minimum height=1em]
\tikzstyle{score} = [draw, minimum width=2em]
\tikzstyle{dep} = [->, looseness=0.4]

\aclfinalcopy
\def\aclpaperid{1820}

\title{Latent Retrieval for Weakly Supervised\\Open Domain Question Answering}

\author{Kenton Lee \quad Ming-Wei Chang \quad Kristina Toutanova \\
 Google Research \\
 Seattle, WA \\
 {\tt \{kentonl,mingweichang,kristout\}@google.com} \\}

\date{}

\begin{document}

\maketitle

\begin{abstract}
Recent work on open domain question answering (QA) assumes strong supervision of the supporting evidence and/or assumes a blackbox information retrieval (IR) system to retrieve evidence candidates. We argue that both are suboptimal, since gold evidence is not always available, and QA is fundamentally different from IR. We show for the first time that it is possible to jointly learn the retriever and reader from question-answer string pairs and without any IR system. In this setting, evidence retrieval from all of Wikipedia is treated as a latent variable. Since this is impractical to learn from scratch, we pre-train the retriever with an \emph{Inverse Cloze Task}. We evaluate on open versions of five QA datasets. On datasets where the questioner already knows the answer, a traditional IR system such as BM25 is sufficient. On datasets where a user is genuinely seeking an answer, we show that learned retrieval is crucial, outperforming BM25 by up to 19 points in exact match.
\end{abstract}

\section{Introduction}
Due to recent advances in reading comprehension systems, there has been a revival of interest in open domain question answering (QA), where the evidence must be retrieved from an open corpus, rather than being given as input. This presents a more realistic scenario for practical applications.

Current approaches require a blackbox information retrieval (IR) system to do much of the heavy lifting, even though it cannot be fine-tuned on the downstream task. In the strongly supervised setting popularized by DrQA~\cite{drqa}, they also assume a reading comprehension model trained on question-answer-evidence triples, such as SQuAD~\cite{squad}. The IR system is used at test time to generate evidence candidates in place of the gold evidence. In the weakly supervised setting, proposed by TriviaQA~\cite{triviaqa}, SearchQA~\cite{searchqa}, and Quasar~\cite{quasar}, the dependency on strong supervision is removed by assuming that the IR system provides noisy gold evidence.

These approaches rely on the IR system to massively reduce the search space and/or reduce spurious ambiguity. However, QA is fundamentally different from IR~\cite{qaretrieval}. Whereas IR is concerned with lexical and semantic matching, questions are \emph{by definition} under-specified and require more language understanding, since users are explicitly looking for unknown information. Instead of being subject to the recall ceiling from blackbox IR systems, we should directly learn to retrieve using question-answering data.

\begin{table*}[t]
\centering
\small
\setlength{\tabcolsep}{9pt}
\scalebox{1.0}{
\begin{tabular}{lccccl}
\toprule 
\multirow{2}{*}{\textbf{Task}} & \multicolumn{2}{c}{\textbf{Training}}& \multicolumn{2}{c}{\textbf{Evaluation}} & \multirow{2}{*}{\textbf{Example}}\\
& \textbf{Evidence} & \textbf{Answer} & \textbf{Evidence} & \textbf{Answer} & \\
\midrule
Reading Comprehension   & given  & span  &  given & string & SQuAD~\cite{squad}\\ 
Open-domain QA \\
~~~~Unsupervised QA & none & none & none & string & GPT-2~\cite{gpt2}\\
~~~~Strongly Supervised QA&given & span  & heuristic & string &  DrQA~\cite{drqa}\\
~~~~Weakly Supervised QA \\
~~~~~~~~Closed Retrieval QA  & heuristic & string & heuristic & string & TriviaQA~\cite{triviaqa}\\
~~~~~~~~\textbf{Open Retrieval QA} &\textbf{learned}& string & \textbf{learned} & string & \textbf{ORQA~(this work)}\\
\bottomrule
\end{tabular}
}
\caption{
Comparison of assumptions made by related tasks, along with references to examples. Heuristic evidence refers to the typical strategy of considering only a closed set of evidence documents from a traditional IR system, which sets a strict upper-bound on task performance. In this work (ORQA), only question-answer string pairs are observed during training, and evidence retrieval is learned in a completely end-to-end manner.
} 
\label{tab:tasks}
\end{table*}

In this work, we introduce the first \emph{Open-Retrieval Question Answering} system (ORQA). ORQA learns to retrieve evidence from an open corpus, and is supervised only by question-answer string pairs. While recent work on improving evidence retrieval has made significant progress~\cite{r3, adaptive, recall, multistep}, they still only rerank a closed evidence set. The main challenge to fully end-to-end learning is that retrieval over the open corpus must be  considered a latent variable that would be impractical to train from scratch. IR systems offer a reasonable but potentially suboptimal starting point.

The key insight of this work is that end-to-end learning \emph{is} possible if we pre-train the retriever with an unsupervised \emph{Inverse Cloze Task} (ICT). In ICT, a sentence is treated as a pseudo-question, and its context is treated as pseudo-evidence. Given a pseudo-question, ICT requires selecting the corresponding pseudo-evidence out of the candidates in a batch. ICT pre-training provides a sufficiently strong initialization such that ORQA, a joint retriever and reader model, can be fine-tuned end-to-end by simply optimizing the marginal log-likelihood of correct answers that were found.

We evaluate ORQA on open versions of five existing QA datasets. On datasets where the question writers already know the answer---SQuAD~\cite{squad} and TriviaQA~\cite{triviaqa}---the retrieval problem resembles traditional IR, and BM25~\cite{bm25} provides state-of-the-art retrieval. On datasets where question writers do not know the answer---Natural Questions~\cite{naturalquestions}, WebQuestions~\cite{webquestions}, and CuratedTrec~\cite{curatedtrec}---we show that learned retrieval is crucial, providing improvements of 6 to 19 points in exact match over BM25.

\section{Overview}
In this section, we introduce notation for open domain QA that is useful for comparing prior work, baselines, and our proposed model.

\subsection{Task}
In open domain question answering, the input $q$ is a question string, and the output $a$ is an answer string. Unlike reading comprehension, the source of evidence is a modeling choice rather than a part of the task definition. We compare the assumptions made by variants of reading comprehension and question answering tasks in Table~\ref{tab:tasks}.

Evaluation is exact match with any of the reference answer strings after minor normalization such as lowercasing, following evaluation scripts from DrQA~\cite{drqa}.

\subsection{Formal Definitions}
We introduce several general definitions of model components that subsume many retrieval-based open domain question answering systems.

Models are defined with respect to an unstructured text corpus that is split into $B$ blocks of evidence texts.
An answer derivation is a pair $(b, s)$, where $1 \le b \le B$ indicates the index of an evidence block and $s$ denotes a span of text within block $b$. The start and end token indices of span $s$ are denoted by \textsc{start}(s) and \textsc{end}(s) respectively.

Models define a scoring function $S(b, s, q)$ indicating the goodness of an answer derivation $(b, s)$ given a question $q$. Typically, this scoring function is decomposed over a retrieval component $S_{\mathit{retr}}(b, q)$  and a reader component $S_{\mathit{read}}(b, s, q)$:
\begin{equation*}
S(b, s, q) = S_{\mathit{retr}}(b, q) + S_{\mathit{read}}(b, s, q)
\end{equation*}
During inference, the model outputs the answer string of the highest scoring derivation:
\begin{equation*}
a^* = \textsc{text}(\operatorname*{argmax}_{b, s}S(b, s, q))
\end{equation*}
where $\textsc{text}(b, s)$ deterministically maps answer derivation $(b, s)$ to an answer string.
A major challenge of any open domain question answering system is handling the scale. In our experiments on the English Wikipedia corpus, we consider over 13 million evidence blocks $b$, each with over 2000 possible answer spans $s$.

\begin{figure*}[t]
\centering
\scalebox{0.6}{
\begin{tikzpicture}[auto, node distance=1.5cm,>=latex']
    \node [bert_q] (bert_q) {$\textsc{BERT}_Q(q)$};
    \node [input, below=of bert_q] (question) {\texttt{[CLS]}What does the zip 
     in zip code stand for?\texttt{[SEP]}};
     
    \node [bert_b, right of=bert_q, yshift=4cm, node distance=8.2cm] (bert_b0) {$\textsc{BERT}_B(0)$};
    \node [input, below=of bert_b0] (block0) {
         \texttt{[CLS]}...The term `ZIP' is
     an acronym for Zone
     Improvement Plan...\texttt{[SEP]}};

    \node [bert_b, below of=bert_b0, node distance=3cm] (bert_b1) {$\textsc{BERT}_B(1)$};
    \node [input, below=of bert_b1] (block1) {
         \texttt{[CLS]}...group of zebras
    are referred to as   a herd or dazzle...\texttt{[SEP]}};

    \node [bert_b, below of= bert_b1, node distance=3cm] (bert_b2) {$\textsc{BERT}_B(2)$};
    \node [input, below=of bert_b2] (block2) {
         \texttt{[CLS]}...ZIPs for other operating systems may     be preceded by...\texttt{[SEP]}};
     
    \node [bert_b, below of= bert_b2, node distance=3cm] (bert_b3) {$\textsc{BERT}_B(...)$};
    \node [input, below=of bert_b3] (block3) {
         ...};
    
    \node [score, left of=bert_b0, node distance=3.7cm](score0) {$S_\mathit{retr}(0, q)$};
    \node [score, left of=bert_b1, node distance=3.7cm](score1) {$S_\mathit{retr}(1, q)$};
    \node [score, left of=bert_b2, node distance=3.7cm](score2) {$S_\mathit{retr}(2, q)$};
    \node [score, left of=bert_b3, node distance=3.7cm](score3) {$S_\mathit{retr}(..., q)$};

    \draw [dep] (bert_q) to [out=0,in=180] (score0);
    \draw [dep] (bert_b0) to [out=180,in=0] (score0);
    
    \draw [dep] (bert_q) to [out=0,in=180] (score1);
    \draw [dep] (bert_b1) to [out=180,in=0] (score1);
    
    \draw [dep] (bert_q) to [out=0,in=180] (score2);
    \draw [dep] (bert_b2) to [out=180,in=0] (score2);
    
    \draw [dep] (bert_q) to [out=0,in=180] (score3);
    \draw [dep] (bert_b3) to [out=180,in=0] (score3);

    \node [bert_r, right of=bert_b0, node distance=6.1cm, yshift=0cm] (bert_r0) {$\textsc{BERT}_R(q, 0)$};
    \node [input, below=of bert_r0] (seq0) {
      \texttt{[CLS]} What does the zip
     in zip code stand for? 
     \texttt{[SEP]}...The term `ZIP' is
     an acronym for Zone
     Improvement Plan...\texttt{[SEP]}};
    
    \node [bert_r, right of=bert_b2, node distance=6.1cm, yshift=0.4cm] (bert_r2) {$\textsc{BERT}_R(q, 2)$};
    \node [input, below=of bert_r2] (seq2) {
      \texttt{[CLS]} What does the zip 
     in zip code stand for? 
     \texttt{[SEP]}...ZIPs for other 
    operating systems may
     be preceded by...\texttt{[SEP]}};
    
    \draw [dep] (block0) to [out=0, in=180, edge node={node [above, yshift=0.2cm] {Top K}}] (seq0);
    \draw [dep] (block2) to [out=0, in=180, edge node={node [above] {Top K}}] (seq2);
    
    \node [score, right of=seq0, node distance=3.5cm, yshift=2.0cm, anchor=west] (span_00) {$S_\mathit{read}(0, \text{``The term''}, q)$};
    \node [score, right of=seq0, node distance=3.5cm, yshift=0.5cm, anchor=west] (span_01) {$S_\mathit{read}(0, \text{``Zone Improvement Plan'', q})$};
    \node [score, right of=seq0, node distance=3.5cm, yshift=-1.0cm, anchor=west] (span_02) {$S_\mathit{read}(0, \text{...}, q)$};
    
    \draw [dep] (bert_r0) to [out=0, in=180, edge node={node [above, sloped] {\scriptsize MLP}}] (span_00);
    \draw [dep] (bert_r0) to [out=0, in=180, edge node={node [above, sloped] {\scriptsize MLP}}] (span_01);
    \draw [dep] (bert_r0) to [out=0, in=180, edge node={node [below, sloped] {\scriptsize MLP}}] (span_02);

    \node [score, right of=seq2, node distance=3.5cm, yshift=2.0cm, anchor=west] (span_20) {$S_\mathit{read}(2, \text{``ZIPs''}, q)$};
    \node [score, right of=seq2, node distance=3.5cm, yshift=0.5cm, anchor=west] (span_21) {$S_\mathit{read}(2, \text{``operating systems'', q})$};
    \node [score, right of=seq2, node distance=3.5cm, yshift=-1.0cm, anchor=west] (span_22) {$S_\mathit{read}(2, \text{...}, q)$};
    
    \draw [dep] (bert_r2) to [out=0, in=180, edge node={node [above, sloped] {\scriptsize MLP}}] (span_20);
    \draw [dep] (bert_r2) to [out=0, in=180, edge node={node [above, sloped] {\scriptsize MLP}}] (span_21);
    \draw [dep] (bert_r2) to [out=0, in=180, edge node={node [below, sloped] {\scriptsize MLP}}] (span_22);
\end{tikzpicture}
}
\caption{Overview of ORQA. A subset of all possible answer derivations given a question $q$ is shown here.  Retrieval scores $S_\mathit{retr}(q, b)$ are computed via inner products between BERT-based encoders.
Top-scoring evidence blocks are jointly encoded with the question, and span representations are scored with a multi-layer perceptron (MLP) to compute $S_\mathit{read}(q, b, s)$. The final joint model score is  $S_\mathit{retr}(q, b)$ + $S_\mathit{read}(q, b, s)$. Unlike previous work using IR systems for candidate proposal, we learn to retrieve from all of Wikipedia directly.}
\label{fig:orqa}
\end{figure*}
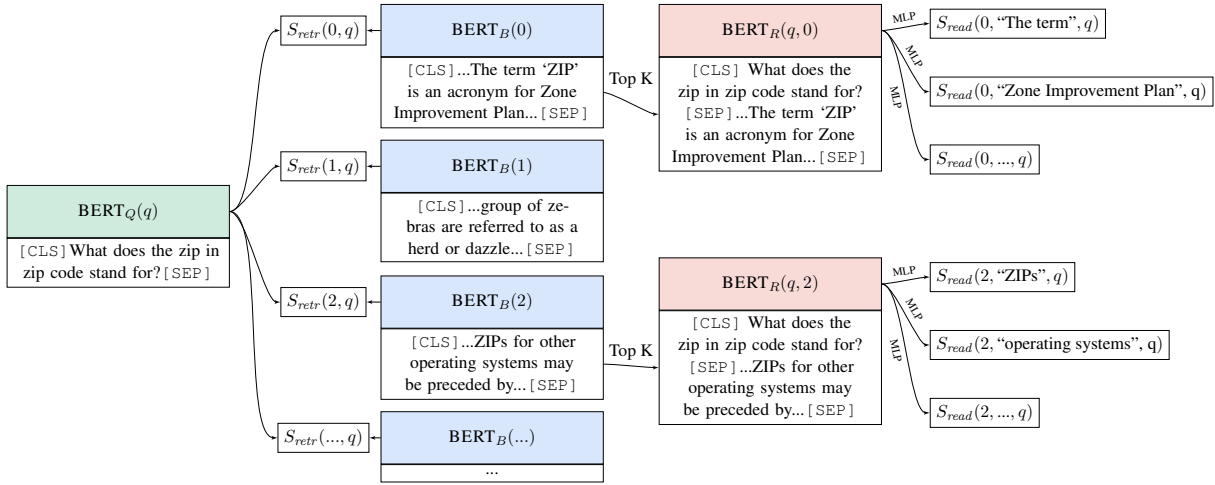
\subsection{Existing Pipelined Models}
In existing retrieval-based open domain question answering systems, a blackbox IR system first chooses a closed set of evidence candidates. For example, the score from the retriever component of DrQA~\cite{drqa} is defined as:
\begin{equation*}
S_{\mathit{retr}}(b, q) = \begin{cases} 
0 & b \in \textsc{top}(k, \textsc{tf-idf}(q, b)) \\ 
-\infty & \text{otherwise}
\end{cases}
\end{equation*}
Most work following DrQA use the same candidates from TF-IDF and focus on reading comprehension or re-ranking. The reading component $S_{\mathit{read}}(b, s, q)$ is learned from gold answer derivations, typically from the SQuAD~\cite{squad} dataset, where the evidence text is given.

In work that is more closely related to our approach, the reader is learned entirely from weak supervision~\cite{triviaqa, quasar, searchqa}. Spurious ambiguities (see Table~\ref{tab:spurious}) are heuristically removed by the retrieval system, and the cleaned results are treated as gold derivations.

\section{Open-Retrieval Question Answering (ORQA)}
\label{sec:model}
We propose an end-to-end model where the retriever and reader components are jointly learned, which we refer to as the Open-Retrieval Question Answering (ORQA) model. An important aspect of ORQA is its expressivity---it is capable of retrieving any text in an open corpus, rather than being limited to the closed set returned by a black-box IR system. An illustration of how ORQA scores answer derivations is presented in Figure~\ref{fig:orqa}. 

Following recent advances in transfer learning, all scoring components are derived from BERT~\cite{bert}, a bidirectional transformer that has been pre-trained on unsupervised language-modeling data. We refer the reader to the original paper for details of the architecture. In this work, the relevant abstraction can be described by the following function:
\begin{equation*}
\textsc{BERT}(x_1, [x_2]) = \{\texttt{CLS}:h_{\texttt{CLS}}, 1:h_1, 2:h_2, ...\}
\end{equation*}
The \textsc{BERT} function takes one or two string inputs ($x_1$ and optionally $x_2$) as arguments. It returns vectors corresponding to representations of the \texttt{CLS} pooling token or the input tokens.

\paragraph{Retriever component}
In order for the retriever to be learnable, we define the retrieval score as the inner product of dense vector representations of the question $q$ and the evidence block $b$.
\begin{align*}
h_\mathit{q} &=  \mathbf{W_q}\textsc{BERT}_Q(q)[\texttt{CLS}] \\
h_\mathit{b} &=  \mathbf{W_b}\textsc{BERT}_B(b)[\texttt{CLS}] \\
S_{\mathit{retr}}(b, q) &= h_\mathit{q} ^\top h_\mathit{b}
\end{align*}
where $\mathbf{W_q}$ and $\mathbf{W_b}$ are matrices that project the \textsc{BERT} output into 128-dimensional vectors.

\paragraph{Reader component}
The reader is a span-based variant of the reading comprehension model proposed in \newcite{bert}:
\begin{align*}
h_\mathit{start} &=  \textsc{BERT}_R(q, b)[\textsc{start}(s)] \\
h_\mathit{end} &=  \textsc{BERT}_R(q, b)[\textsc{end}(s)] \\
S_{\mathit{read}}(b, s, q) &= \textsc{MLP}([h_\mathit{start}; h_\mathit{end}])
\end{align*}
Following \newcite{rasor}, a span is represented by the concatenation of its end points, which is scored by a multi-layer perceptron to enable start/end interaction.

\begin{table}[t!]
\small
\centering
\begin{tabularx}{\linewidth}{L{2.4cm}ll}
\toprule
\multirow{2}{*}{\textbf{Example}} & \textbf{Supportive} & \textbf{Spurious} \\
&\textbf{Evidence} & \textbf{Ambiguity}\\
\midrule
\textbf{Q}: Who is credited with developing the XY coordinate plane?
& \multirow{2}{2.2cm}{...invention of Cartesian coordinates by \textbf{René Descartes} revolutionized...}
& \multirow{2}{2.1cm}{...\textbf{René Descartes} was born in La Haye en Touraine, France...}\\
 \textbf{A}: René Descartes & & \\
 \cmidrule{1-3}
\textbf{Q}: How many districts are in the state of Alabama?
& \multirow{2}{2.2cm}{...Alabama is currently divided into \textbf{seven} congressional districts, each represented by ...}
& \multirow{2}{2.1cm}{...Alabama is one of \textbf{seven} states that levy a tax on food at the same rate as other goods...}\\
 \textbf{A}: seven & &\\
 & &\\
 & &\\
\bottomrule
\end{tabularx}
\caption{Examples of spurious ambiguities arising from the use of weak supervision. Good evidence retrieval is needed to generate a meaningful learning signal.}
\label{tab:spurious}
\end{table}

\paragraph{Inference \& Learning Challenges}
The model described above is conceptually simple. However, inference and learning are challenging since (1) an open evidence corpus presents an enormous search space (over 13 million evidence blocks), and (2) how to navigate this space is entirely latent, so standard teacher-forcing approaches do not apply. Latent-variable methods are also difficult to apply naively due to the large number of spuriously ambiguous derivations. For example, as shown in Table~\ref{tab:spurious}, many irrelevant passages in Wikipedia would contain the answer string ``seven.''

We address these challenges by carefully initializing the retriever with unsupervised pre-training (Section~\ref{sec:ict}). The pre-trained retriever allows us to (1) pre-encode all evidence blocks from Wikipedia, enabling dynamic yet fast top-k retrieval during fine-tuning (Section~\ref{sec:inference}), and (2) bias the retrieval away from spurious ambiguities and towards supportive evidence (Section~\ref{sec:learning}).

\section{Inverse Cloze Task}
\label{sec:ict}
The goal of our proposed pre-training procedure is for the retriever to solve an unsupervised task that closely resembles evidence retrieval for QA.

Intuitively, useful evidence typically discusses entities, events, and relations from the question. It also contains extra information (the answer) that is \emph{not} present in the question. An unsupervised analog of a question-evidence pair is a sentence-context pair---the context of a sentence is semantically relevant and can be used to infer information missing from the sentence.

\begin{figure}[t!]
\centering
\scalebox{0.57}{
\begin{tikzpicture}[auto, node distance=1.5cm,>=latex']
    \node [bert_q] (bert_q) {$\textsc{BERT}_Q(q)$};
    \node [input, below=of bert_q] (question) {\texttt{[CLS]}They are generally slower than horses, but their great stamina helps them outrun predators.\texttt{[SEP]}};
     
    \node [bert_b, right of=bert_q, yshift=4cm, node distance=8.2cm] (bert_b0) {$\textsc{BERT}_B(0)$};
    \node [input, below=of bert_b0] (block0) {
         \texttt{[CLS]}...Zebras have four gaits: walk, trot, canter and gallop. When chased, a zebra will zig-zag from side to side... 
...\texttt{[SEP]}};

    \node [bert_b, below of=bert_b0, node distance=4cm] (bert_b1) {$\textsc{BERT}_B(1)$};
    \node [input, below=of bert_b1] (block1) { \texttt{[CLS]}...Gagarin was further selected for an elite training group known as the Sochi Six...\texttt{[SEP]}};

    \node [bert_b, below of= bert_b1, node distance=3.5cm] (bert_b2) {$\textsc{BERT}_B(...)$};
    \node [input, below=of bert_b2] (block2) {...};
    
    \node [score, left of=bert_b0, node distance=3.7cm](score0) {$S_\mathit{retr}(0, q)$};
    \node [score, left of=bert_b1, node distance=3.7cm](score1) {$S_\mathit{retr}(1, q)$};
    \node [score, left of=bert_b2, node distance=3.7cm](score2) {$S_\mathit{retr}(..., q)$};
    
    \draw [dep] (bert_q) to [out=0,in=180] (score0);
    \draw [dep] (bert_b0) to [out=180,in=0] (score0);
    
    \draw [dep] (bert_q) to [out=0,in=180] (score1);
    \draw [dep] (bert_b1) to [out=180,in=0] (score1);
    
    \draw [dep] (bert_q) to [out=0,in=180] (score2);
    \draw [dep] (bert_b2) to [out=180,in=0] (score2);
\end{tikzpicture}
}
\caption{Example of the Inverse Cloze Task (ICT), used for retrieval pre-training. A random sentence (pseudo-query) and its context (pseudo evidence text) are derived from the text snippet: 
\textit{``...Zebras have four gaits: walk, trot, canter and gallop.
\textbf{They are generally slower than horses, but their great stamina helps them outrun predators}.
When chased, a zebra will zig-zag from side to side...''} The objective is to select the true context among candidates in the batch.}
\label{fig:ict}
\end{figure}
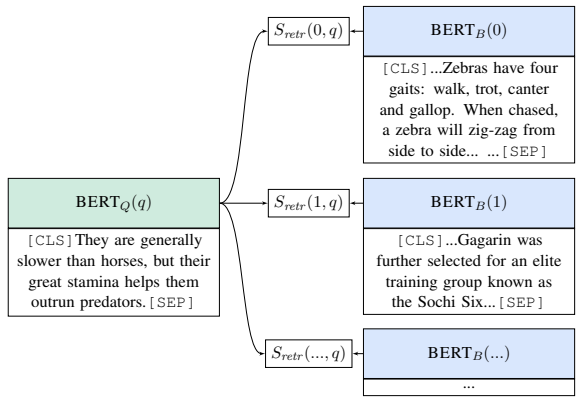

Following this intuition, we propose to pre-train our retrieval module with an Inverse Cloze Task (ICT). In the standard Cloze task~\cite{cloze}, the goal is to predict masked-out text based on its context. ICT instead requires predicting the inverse---given a sentence, predict its context (see Figure~\ref{fig:ict}). We use a discriminative objective that is analogous to downstream retrieval:
\begin{equation*}
P_\textsc{ict}(b | q) = \frac{\exp(S_{\mathit{retr}}(b, q))}{\displaystyle\sum_{b' \in \textsc{batch}}\exp(S_{\mathit{retr}}(b', q))}
\end{equation*}
where $q$ is a random sentence that is treated as a pseudo-question, $b$ is the text surrounding $q$, and $\textsc{batch}$ is the set of evidence blocks in the batch that are used as sampled negatives.

An important aspect of ICT is that it requires learning more than word matching features, since the pseudo-question is not present in the evidence. For example, the pseudo-question in Figure~\ref{fig:ict} never explicitly mentions ``Zebras'', but the retriever must still be able to select the context that discusses Zebras. Being able to infer the semantics from under-specified language is what sets QA apart from traditional IR.

However, we also do not want to dissuade the retriever from learning to perform word matching---lexical overlap is ultimately a very useful feature for retrieval. Therefore, we only remove the sentence from its context in 90\% of the examples, encouraging the model to learn both abstract representations when needed and low-level word matching features when available.
\\\\
ICT pre-training accomplishes two main goals:\begin{enumerate}
    \item Despite the mismatch between sentences during pre-training and questions during fine-tuning, we expect \emph{zero-shot} evidence retrieval performance to be sufficient for bootstrapping the latent-variable learning.
    \item There is no such mismatch between pre-trained evidence blocks and downstream evidence blocks. We can expect the  block encoder $\textsc{BERT}_B(b)$ to work well without further training. Only the question encoder needs to be fine-tuned on downstream data.
\end{enumerate}
As we will see in the following section, these two properties are crucial for enabling computationally feasible inference and end-to-end learning.

\begin{table*}[t!]
\small
\centering
\begin{tabularx}{\linewidth}{L{2.3cm}P{0.7cm}P{0.7cm}P{0.7cm}p{6cm}p{3.3cm}}
\toprule
\textbf{Dataset} & \textbf{Train} & \textbf{Dev} & \textbf{Test} & \textbf{Example Question} & \textbf{Example Answer} \\
\midrule
Natural Questions & 79168 & 8757 & 3610 & \small What does the zip in zip code stand for? & \small Zone Improvement Plan\\ 
WebQuestions & 3417 & 361 & 2032 & \small What airport is closer to downtown Houston? & \small William P. Hobby Airport\\
CuratedTrec & 1353 & 133 & 694 & \small What metal has the highest melting point?& \small Tungsten\\ 
TriviaQA & 78785 & 8837 & 11313 & \small What did L. Fran Baum, author of The Wonderful Wizard of Oz, call his home in Hollywood? & \small Ozcot\\
SQuAD & 78713 & 8886 & 10570 & \small Other than the Automobile Club of Southern California, what other AAA Auto Club chose to simplify the divide? & \small California State Automobile Association\\
\bottomrule
\end{tabularx}
\caption{Statistics and examples for the datasets that we evaluate on. There are slightly differences from the original datasets as described in Section~\ref{sec:datasets}, since not all of them were intended to be used in the open setting.}
\label{tab:datasets}
\end{table*}

\section{Inference}
\label{sec:inference}
Since fixed block encoders already provide a useful representation for retrieval, we can pre-compute all block encodings in the evidence corpus. As a result, the enormous set of evidence blocks does not need to be re-encoded while fine-tuning, and it can be pre-compiled into an index for fast maximum inner product search using existing tools such as Locality Sensitive Hashing.

With the pre-compiled index, inference follows a standard beam-search procedure. We retrieve the top-$k$ evidence blocks and only compute the expensive reader scores for those $k$ blocks. While we only consider the top-$k$ evidence blocks during a single inference step, this set dynamically changes during training since the question encoder is fine-tuned according to the weakly supervised QA data, as discussed in the following section.

\section{Learning}
\label{sec:learning}
Learning is relatively straightforward, since ICT should provide non-trivial zero-shot retrieval. We first define a distribution over answer derivations:
\begin{equation*}
P(b, s | q) = \frac{\exp(S(b, s, q))}{\displaystyle\sum_{b'\in \textsc{top}(k)}\displaystyle\sum_{s' \in b'}\exp(S(b', s', q))}
\end{equation*}
where $\textsc{top}(k)$ denotes the top $k$ retrieved blocks based on $S_{\mathit{retr}}$. We use $k = 5$ in our experiments.

Given a gold answer string $a$, we find all (possibly spuriously) correct derivations in the beam, and optimize their marginal log-likelihood:
\begin{equation*}
L_{\text{full}}(q, a) = -\log\hspace{-0.2cm}\displaystyle\sum_{b\in \textsc{top}(k)}\displaystyle\sum_{s \in b,~ a=\textsc{text}(s)}\hspace{-0.6cm}P'(b, s | q)
\end{equation*}
where $a = \textsc{text}(s)$ indicates whether the answer string $a$ matches exactly the span $s$.

To encourage more aggressive learning, we also include an early update, where we consider a larger set of $c$ evidence blocks but only update the retrieval score, which is cheap to compute:
\begin{align*}
P_\text{early}(b | q) &= \frac{\exp(S_{\mathit{retr}}(b, q))}{\displaystyle\sum_{b'\in \textsc{top}(c)}\exp(S_{\mathit{retr}}(b', q))}\\
L_{\text{early}}(q, a) &= -\log\hspace{-0.7cm}\sum_{b \in \textsc{top}(c),~a \in \textsc{text}(b)}\hspace{-0.8cm} P_\text{early}(b | q)
\end{align*}
where $a \in \textsc{text}(b)$ indicates whether answer string $a$ appears in evidence block $b$. We use $c = 5000$ in our experiments.

The final loss includes both updates:
\begin{equation*}
    L(q, a) = L_{\text{early}}(q, a) + L_{\text{full}}(q, a)
\end{equation*}
If no matching answers are found at all, then the example is discarded. While we would expect almost all examples to be discarded with random initialization, we discard less than 10\% of examples in practice due to ICT pre-training.

As previously mentioned, we fine-tune all parameters except those in the evidence block encoder. Since the query encoder is trainable, the model can potentially learn to retrieve any evidence block. This expressivity is a crucial difference from blackbox IR systems, where recall can only be improved by retrieving more evidence.

\section{Experimental Setup}
\subsection{Open Domain QA Datasets}
\label{sec:datasets}
We train and evaluate on data from 5 existing question answering or reading comprehension datasets. Not all of them are intended as open domain QA datasets in their original form, so we convert them to open formats, following DrQA~\cite{drqa}.
Each example in the open version of the datasets consists of a single question string and a set of reference answer strings.

\paragraph{Natural Questions} contains question from aggregated queries to Google Search~\cite{naturalquestions}. To gather an open version of this dataset, we only keep questions with short answers and discard the given evidence document. Answers with many tokens often resemble extractive snippets rather than canonical answers, so we discard answers with more than 5 tokens.
\paragraph{WebQuestions} contains questions that were sampled from the Google Suggest API~\cite{webquestions}. The answers are annotated with respect to Freebase, but we only keep the string representation of the entities.
\paragraph{CuratedTrec} is a corpus of question-answer pairs derived from TREC QA data curated by \newcite{curatedtrec}. The questions come from various sources of real queries, such as MSNSearch or AskJeeves logs, where the question askers do not observe any evidence documents~\cite{trec01}.
\paragraph{TriviaQA} is a collection of trivia question-answer pairs that were scraped from the web ~\cite{triviaqa}. We use their unfiltered set and discard their distantly supervised evidence.
\paragraph{SQuAD} was designed to be a reading comprehension dataset rather than an open domain QA dataset~\cite{squad}. Answer spans were selected from a Wikipedia paragraph, and the questions were written by annotators who were instructed to ask questions that are answered by a given answer in a given context.
\\\\
On datasets where a development set does not exist, we randomly hold out 10\% of the training data for development. On datasets where the test set is hidden, we also randomly hold out 10\% of the training data for development, and use the original development set for testing (following DrQA). A summary of dataset statistics and examples are shown in Table~\ref{tab:datasets}.

\subsection{Dataset Biases}
Evaluating on this diverse set of question-answer pairs is crucial, because all existing datasets have inherent biases that are problematic for open domain QA systems with learned retrieval. These biases are summarized in Table~\ref{tab:biases}.

In the Natural Questions, WebQuestions, and CuratedTrec, the question askers do not already know the answer. This accurately reflects a distribution of genuine information-seeking questions. However, annotators must separately find correct answers, which requires assistance from automatic tools and can introduce a moderate bias towards results from the tool.

In TriviaQA and SQuAD, automatic tools are not needed since the questions are written with known answers in mind. However, this introduces another set of biases that are arguably more problematic. Question writing is not motivated by an information need. This often results in many hints in the question that would not be present in naturally occurring questions, as shown in the examples in Table~\ref{tab:datasets}. This is particularly problematic for SQuAD, where the question askers are also prompted with a specific piece of evidence for the answer, leading to artificially large lexical overlap between the question and evidence.

Note that these are simply properties of the datasets rather than actionable criticisms---such data collection methods are necessary to scale up, and it is unclear how one could collect a truly unbiased dataset without impractical costs.

\subsection{Implementation Details}
We mainly evaluate in the setting where only question-answer string pairs are available for supervision. See Section~\ref{sec:analysis} for head-to-head comparisons with the DrQA setting that uses the same evidence corpus and the same type of supervision.

\paragraph{Evidence Corpus}
We use the English Wikipedia snapshot from December 20, 2018 as the evidence corpus.\footnote{We deviate from DrQA's 2016 Wikipedia evidence corpus because the original snapshot is no longer publicly available. The 12-20-2018 snapshot is available at \scriptsize{\url{https://archive.org/download/enwiki-20181220}}.}
The corpus is greedily split into chunks of at most 288 wordpieces based on BERT's tokenizer, while preserving sentence boundaries. This results in just over 13 million evidence blocks. The title of the document is included in the block encoder.

\paragraph{Hyperparameters}
In all uses of BERT (both the retriever and reader), we initialize from the uncased base model, which consists of 12 transformer layers with a hidden size of 768.

As mentioned in Section~\ref{sec:model}, the retrieval representations, $h_\mathit{q}$ and $h_\mathit{b}$, have 128 dimensions. The small hidden size was chosen so that the final QA model can comfortably run on a single machine. We use the default optimizer from BERT.

When pre-training the retriever with ICT, we use a learning rate of $10^{-4}$ and a batch size of 4096 on Google Cloud TPUs for 100k steps. When fine-tuning, we use a learning rate of $10^{-5}$ and a batch size of 1 on a single machine with a 12GB GPU. Answer spans are limited to 10 tokens. We perform 2 epochs of fine-tuning for the larger datasets (Natural Questions, TriviaQA, and SQuAD), and 20 epochs for the smaller datasets (WebQuestions and CuratedTrec).

\begin{table}[t!]
\small
\centering
\begin{tabularx}{\linewidth}{p{2.3cm}ccc}
\toprule
\textbf{Dataset} & \textbf{Question} & \textbf{Question} & \textbf{Tool-} \\
&  \textbf{writer} & \textbf{writer} & \textbf{assisted} \\
&  \textbf{knows} & \textbf{knows} & \textbf{answer} \\
&  \textbf{answer} & \textbf{evidence} & \\
\midrule
Natural Questions & & & \yesmark\\
WebQuestions & & & \yesmark\\
CuratedTrec & & & \yesmark\\
\cmidrule{2-4}
TriviaQA & \yesmark & &\\
SQuAD & \yesmark & \yesmark &\\
\bottomrule
\end{tabularx}
\caption{A breakdown of biases in existing QA datasets. These biases are associated with either the question or the answer.}
\label{tab:biases}
\end{table}

\section{Main Results}
\subsection{Baselines}
We compare against other retrieval methods by using alternate retrieval scores $S_{\mathit{retr}}(b, q)$, but with the same reader.

\paragraph{BM25} A de-facto state-of-the-art unsupervised retrieval method is BM25~\cite{bm25}. It has been shown to be robust for both traditional information retrieval tasks, and evidence retrieval for question answering~\cite{anserini}.\footnote{We also include the title, which was slightly beneficial.} Since BM25 is not trainable, the retrieved evidence considered during fine-tuning is static. Inspired by BERTserini~\cite{bertserini}, the final score is a learned weighted sum of the BM25 and reader score. Our implementation is based on Lucene.\footnote{\scriptsize \url{https://lucene.apache.org/}}

\paragraph{Language Models} While unsupervised neural retrieval is notoriously difficult to improve over traditional IR~\cite{hype}, we include them as baselines for comparison. We experiment with unsupervised pooled representations from neural language models (LM), which has been shown to be state-of-the-art unsupervised representations~\cite{elmopool}. We compare with two widely-used 128-dimensional representations: (1) \textsc{NNLM}, context-\emph{independent} embeddings from a feed-forward LMs~\cite{nnlm},\footnote{\scriptsize\url{https://tfhub.dev/google/nnlm-en-dim128/1}} and (2) \textsc{ELMo}~(small), a context-\emph{dependent} bidirectional LSTM~\cite{elmo}.\footnote{\scriptsize\url{https://allennlp.org/elmo}}

As with ICT, we use the alternate encoders to pre-compute the encoded evidence blocks $h_b$ and to initialize the question encoding $h_q$, which is fine-tuned. Based on existing IR literature and the intuition that LMs do not explicitly optimize for retrieval, we do not expect these to be strong baselines, but they demonstrate the difficulty of encoding blocks of text into 128 dimensions.

\subsection{Results}
\begin{table}[t!]
\small
\centering
\begin{tabularx}{\linewidth}{p{0cm}p{1.9cm}P{0.8cm}P{0.8cm}P{0.8cm}P{0.8cm}}
\toprule
&\multirow{2}{*}{\textbf{Model}}  & \textsc{BM25} & \textsc{NNLM} &  \textsc{ELMo} & \multirow{2}{*}{\textsc{ORQA}}\\
&& \textsc{+BERT} & \textsc{+BERT} &  \textsc{+BERT} \\
\midrule
\parbox[t]{2mm}{\multirow{5}{*}{\rotatebox[origin=c]{90}{Dev}}}&\text{Natural Questions} & 24.8 & 3.2 & 3.6  & \textbf{31.3}\\
&\text{WebQuestions}      & 20.8 & 9.1 & 17.7 & \textbf{38.5}\\
&\text{CuratedTrec}       & 27.1 & 6.0 & 8.3  & \textbf{36.8}\\
\cmidrule{3-6}
&\text{TriviaQA}          & \textbf{47.2} & 7.3 & 6.0  & 45.1\\
&\text{SQuAD}             & \textbf{28.1} & 2.8 & 1.9  & 26.5\\
\midrule
\parbox[t]{2mm}{\multirow{5}{*}{\rotatebox[origin=c]{90}{Test}}}&\text{Natural Questions} &  26.5 & 4.0 & 4.7  & \textbf{33.3}\\
&\text{WebQuestions}      &  17.7 & 7.3 & 15.6 & \textbf{36.4}\\
&\text{CuratedTrec}       &  21.3 & 4.5 & 6.8  & \textbf{30.1}\\ 
\cmidrule{3-6}
&\text{TriviaQA}          &  \textbf{47.1} & 7.1 & 5.7  & 45.0\\
&\text{SQuAD}             &  \textbf{33.2} & 3.2 & 2.3  & 20.2\\
\bottomrule
\end{tabularx}
\caption{\textbf{Main results}: End-to-end exact match for open-domain question answering from question-answer pairs only. Datasets where question askers know the answer behave differently from datasets where they do not.}
\label{tab:results}
\end{table}
\label{sec:results}
The main results are show in Table~\ref{tab:results}. The first result to note is that BM25 is a powerful retrieval system. Word matching is important, and dense vector representations derived from language models do not readily capture this.

We also show that on questions that were derived from real users who are seeking information (Natural Questions, WebQuestions, and CuratedTrec), our ICT pre-trained retriever outperforms BM25 by a large marge---6 to 19 points in exact match depending on the dataset.

However, in datasets where the question askers already know the answer, i.e. SQuAD and TriviaQA, the retrieval problem resembles traditional IR. In this setting, a highly compressed 128-dimensional vector cannot match BM25's ability to precisely represent every word in the evidence.

The notable drop between development and test accuracy for SQuAD is a reflection of an artifact in the dataset---its 100k questions are derived from only 536 documents. Therefore, good retrieval targets are highly correlated between training examples, violating the IID assumption, and making it unsuitable for learned retrieval. We strongly suggest that those who are interested in end-to-end open-domain QA models no longer train and evaluate with SQuAD for this reason.

\section{Analysis}
\label{sec:analysis}
\begin{table}[t!]
\small
\centering
\begin{tabularx}{\linewidth}{L{2.8cm}P{2cm}N}
\toprule
\multirow{2}{*}{\textbf{Model}} & \textbf{Evidence} &\multirow{2}{*}{\textbf{SQuAD}}\\ 
& \textbf{Retrieved} &\\
\midrule
\textsc{DrQA} &  5 documents & 27.1 \\
\textsc{DrQA (DS)} &  5 documents & 28.4 \\
\textsc{DrQA (DS + MTL)} &  5 documents & 29.8 \\ 
\cmidrule{1-3}
\textsc{BERTserini} & 5 documents & 19.1 \\ 
\textsc{BERTserini} & 29 paragraphs & 36.6 \\
\textsc{BERTserini} & 100 paragraphs & 38.6 \\ 
\cmidrule{1-3}
\textsc{BM25 + BERT}  & \multirow{2}{*}{5 blocks} & \multirow{2}{*}{34.7}  \\ 
 (gold deriv.) & &\\
\bottomrule
\end{tabularx}
\caption{\textbf{Analysis}: Results comparable to previous work in the strongly supervised setting, where models have access to gold derivations from SQuAD. Different systems segment Wikipedia differently. There are 5.1M documents, 29.5M paragraphs, and 12.1M blocks in the December 12, 2016 Wikipedia snapshot.}
\label{tab:drqa}
\end{table}
\subsection{Strongly supervised comparison}
To verify that our BM25 baseline is indeed state of the art, we also provide direct comparisons with DrQA's setup, where systems have access to gold answer derivations from SQuAD~\cite{squad}. While many systems have been proposed following DrQA's original setting, we compare only to the original system and the best system that we are aware of---BERTserini~\cite{bertserini}.

DrQA's reader is DocReader~\cite{drqa}, and they use TF-IDF to retrieve the top $k$ documents. They also include distant supervision based on TF-IDF retrieval. BERTserini's reader is derived from base BERT (much like our reader), and they use BM25 to retrieve the top $k$ paragraphs (much like our BM25 baseline). A major difference is that BERTserini uses true paragraphs from Wikipedia rather than arbitrary blocks, resulting in more evidence blocks due to uneven lengths.

For fair comparison with these strongly supervised systems, we pre-train the reader on SQuAD data.\footnote{We use DrQA's December 12, 2016 snapshot of Wikipedia for an apples-to-apples comparison.} In Table~\ref{tab:drqa}, our BM25 baseline, which retrieves 5 evidence blocks, greatly outperforms 5-document \mbox{BERTserini} and is close to 29-paragraph \mbox{BERTserini}.

\begin{figure}[t!]
\begin{tikzpicture}

 \begin{axis}[
   width=0.95\columnwidth,
   height=0.5\columnwidth,
   legend cell align=left,
   legend style={at={(0, 0)},anchor=south west,font=\small},
   mark options={mark size=2},
   font=\small,
   xmin=0, xmax=1,
   ymin=15, ymax=35,
   ytick={20, 25, 30, 35},
   xtick={0, 0.1, 0.2, 0.3, 0.4, 0.5, 0.6, 0.7, 0.8, 0.9, 1.0},
   ymajorgrids=true,
   xmajorgrids=true,
   xlabel style={yshift=0.5ex,},
   xlabel=ICT masking rate,
   ylabel style={align=center},
   ylabel=Natural Questions\\Exact Match,
   ylabel style={yshift=-0.5ex,}]
    \addplot[mark=o, g-blue, line width=1pt] plot coordinates {
      (0, 25.0)
      (0.1, 26.8)
      (0.2, 27.8)
      (0.5, 29.3)
      (0.8, 29.8)
      (0.9, 31.3)
      (0.95, 30.2)
      (1.0, 21.7)
    };
    \addlegendentry{ORQA}
    \addplot[g-red, line width=1pt, dashed] plot coordinates {
      (0.0, 24.8)
      (1.0, 24.8)
    };
    \addlegendentry{BM25 + BERT}
\end{axis}
\end{tikzpicture}
\caption{\textbf{Analysis}: Performance on our open version of the Natural Questions dev set with various masking rates for the ICT pre-training. Too much masking prevents the model from learning to exploit exact n-gram overlap. Too little masking makes language understanding unnecessary.}
\label{tab:mask}
\end{figure}
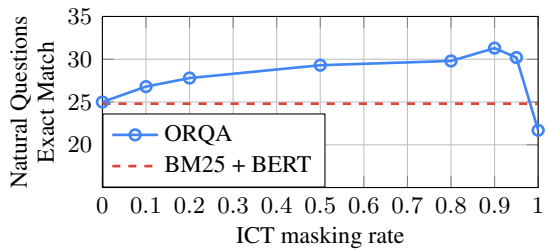
\begin{table*}[t!]
\small
\centering
\begin{tabularx}{\linewidth}{L{2.9cm}ll}
\toprule
\textbf{Example}& \textbf{ORQA} & \textbf{BM25 + BERT} \\
\midrule
\textbf{Q}: what is the new orleans saints symbol called
& \multirow{2}{6cm}{...The team's primary colors are old gold and black; their logo is a simplified \textbf{fleur-de-lis}. They played their home games in Tulane Stadium through the 1974 NFL season....}
& \multirow{2}{6cm}{...the \textbf{SkyDome} was owned by Sportsco at the time... the sale of the New Orleans Saints with team owner Tom Benson... the Saints became a symbol for that community...} \\
\textbf{A}:  fleur-de-lis  & & \\
 \cmidrule{1-3}
\textbf{Q}: how many senators per state in the us 
& \multirow{2}{6cm}{...powers of the Senate are established in Article One of the U.S. Constitution. Each U.S. state is represented by \textbf{two} senators...}
& \multirow{2}{6cm}{...The Georgia Constitution mandates a maximum of \textbf{56} senators, elected from single-member districts...} \\
\textbf{A}:  two & & \\
\cmidrule{1-3}
\textbf{Q}: when was germany given a permanent seat on the council of the league of nations
& \multirow{2}{6cm}{...Under the Weimar Republic, Germany (in fact the ``Deutsches Reich'' or German Empire) was admitted to the League of Nations through a resolution passed on September 8 \textbf{1926}. An additional 15 countries joined later...}
& \multirow{2}{6cm}{...the accession of the German Democratic Republic to the Federal Republic of Germany, it was effective on \textbf{3 October 1990}...Germany has been elected as a non-permanent member of the United Nations Security Council...} \\
\textbf{A}: 1926 & & \\
\cmidrule{1-3}
\textbf{Q}: when was diary of a wimpy kid double down published
& \multirow{2}{6cm}{...``Diary of a Wimpy Kid'' first appeared on FunBrain in 2004, where it was read 20 million times. The abridged hardcover adaptation was released on \textbf{April 1, 2007}...}
& \multirow{2}{6cm}{Diary of a Wimpy Kid: Double Down is the eleventh book in the "Diary of a Wimpy Kid" series by Jeff Kinney... The book was published on \textbf{November 1, 2016}...} \\
\textbf{A}:  November 1, 2016  & & \\
\bottomrule
\end{tabularx}
\caption{\textbf{Analysis}: Example predictions on our open version of the Natural Questions dev set. We show the highest scoring derivation, consisting of the evidence block and the predicted answer in bold. ORQA is more robust at separating semantically distinct text that have high lexical overlap. However, the limitation of the 128-dimensional vectors is that extremely specific concepts are less precisely represented.}
\label{tab:analysis}
\end{table*}
\subsection{Masking Rate in the Inverse Cloze Task}
The pseudo-query is masked from the evidence block 90\% of the time, motivated by intuition in Section~\ref{sec:ict}. We empirically verify our intuitions in Figure~\ref{tab:mask} by varying the masking rate, and comparing results on our open version of the Natural Questions development set.

If we always mask the pseudo-query, the retriever never learns that n-gram overlap is a powerful retrieval signal, losing almost 10 points in end-to-end performance. If we never mask the pseudo-query, the problem is reduced to memorization and does not generalize well to question answering. The latter loses 6 points in end-to-end performance, which---perhaps not surprisingly---produces near-identical results to BM25.

\subsection{Example Predictions}
For a more intuitive understanding of the improvements from ORQA, we compare its predictions with baseline predictions in Table~\ref{tab:analysis}. We find that ORQA is more robust at separating semantically distinct text with high lexical overlap, as shown in the first three examples. However, it is expected that there are limits to how much information can be compressed into 128-dimensional vectors. The last example shows that ORQA has trouble precisely representing extremely specific concepts that sparse representations can cleanly separate. These errors indicate that a hybrid approach would be promising future work.

\section{Related Work}
Recent progress has been made towards improving evidence retrieval~\cite{r3, adaptive, recall, multistep} by learning to aggregate from multiple retrieval steps. They re-rank evidence candidates from a closed set, and we aim to integrate these complementary approaches in future work.

Our approach is also reminiscent of weakly supervised semantic parsing~\cite{driving, compositional, instructional, openqa, webquestions, scaling}, with which we share similar challenges---(1) inference and learning are tightly coupled, (2) latent derivations must be discovered, and (3) strong inductive biases are needed to find positive learning signal while avoiding spurious ambiguities.

While we motivate ICT from first principles as an unsupervised proxy for evidence retrieval, it is closely related to existing representation learning literature. ICT can be considered a generalization of the skip-gram objective~\cite{skipgram}, with a coarser granularity, deep architecture, and in-batch negative sampling from ~\newcite{quickthoughts}.

Consulting external evidence sources with latent retrieval has also been explored in information extraction~\cite{ie_external}. In comparison, we are able to learn a much more expressive retriever due to the strong inductive biases from ICT pre-training.

\section{Conclusion}
We presented ORQA, the first open domain question answering system where the retriever and reader are jointly learned end-to-end using only question-answer pairs and without any IR system. This is made possible by pre-training the retriever using an Inverse Cloze Task (ICT). Experiments show that learning to retrieve is crucial when the questions reflect an information need, i.e. the question writers do not already know the answer.

\section*{Acknowledgements}
We thank the Google AI Language Team for valuable suggestions and feedback.

\bibliography{main}

\begin{thebibliography}{33}
\expandafter\ifx\csname natexlab\endcsname\relax\def\natexlab#1{#1}\fi

\bibitem[{Artzi and Zettlemoyer(2013)}]{instructional}
Yoav Artzi and Luke Zettlemoyer. 2013.
\newblock Weakly supervised learning of semantic parsers for mapping
  instructions to actions.
\newblock \emph{Transactions of the Association for Computational Linguistics},
  1(1):49--62.

\bibitem[{Baudis and Sediv{\'y}(2015)}]{curatedtrec}
Petr Baudis and Jan Sediv{\'y}. 2015.
\newblock Modeling of the question answering task in the yodaqa system.
\newblock In \emph{CLEF}.

\bibitem[{Bengio et~al.(2003)Bengio, Ducharme, Vincent, and Jauvin}]{nnlm}
Yoshua Bengio, R{\'e}jean Ducharme, Pascal Vincent, and Christian Jauvin. 2003.
\newblock A neural probabilistic language model.
\newblock \emph{Journal of machine learning research}, 3(Feb):1137--1155.

\bibitem[{Berant et~al.(2013)Berant, Chou, Frostig, and Liang}]{webquestions}
Jonathan Berant, Andrew Chou, Roy Frostig, and Percy Liang. 2013.
\newblock Semantic parsing on freebase from question-answer pairs.
\newblock In \emph{Proceedings of the 2013 Conference on Empirical Methods in
  Natural Language Processing}, pages 1533--1544.

\bibitem[{Chen et~al.(2017)Chen, Fisch, Weston, and Bordes}]{drqa}
Danqi Chen, Adam Fisch, Jason Weston, and Antoine Bordes. 2017.
\newblock Reading wikipedia to answer open-domain questions.
\newblock In \emph{Proceedings of the 55th Annual Meeting of the Association
  for Computational Linguistics (Volume 1: Long Papers)}, volume~1, pages
  1870--1879.

\bibitem[{Clarke et~al.(2010)Clarke, Goldwasser, Chang, and Roth}]{driving}
James Clarke, Dan Goldwasser, Ming-Wei Chang, and Dan Roth. 2010.
\newblock Driving semantic parsing from the world's response.
\newblock In \emph{Proceedings of the fourteenth conference on computational
  natural language learning}, pages 18--27. Association for Computational
  Linguistics.

\bibitem[{Das et~al.(2019)Das, Dhuliawala, Zaheer, and McCallum}]{multistep}
Rajarshi Das, Shehzaad Dhuliawala, Manzil Zaheer, and Andrew McCallum. 2019.
\newblock \href {https://openreview.net/forum?id=HkfPSh05K7} {Multi-step
  retriever-reader interaction for scalable open-domain question answering}.
\newblock In \emph{International Conference on Learning Representations}.

\bibitem[{Devlin et~al.(2018)Devlin, Chang, Lee, and Toutanova}]{bert}
Jacob Devlin, Ming-Wei Chang, Kenton Lee, and Kristina Toutanova. 2018.
\newblock Bert: Pre-training of deep bidirectional transformers for language
  understanding.
\newblock \emph{arXiv preprint arXiv:1810.04805}.

\bibitem[{Dhingra et~al.(2017)Dhingra, Mazaitis, and Cohen}]{quasar}
Bhuwan Dhingra, Kathryn Mazaitis, and William~W Cohen. 2017.
\newblock Quasar: Datasets for question answering by search and reading.
\newblock \emph{arXiv preprint arXiv:1707.03904}.

\bibitem[{Dunn et~al.(2017)Dunn, Sagun, Higgins, Guney, Cirik, and
  Cho}]{searchqa}
Matthew Dunn, Levent Sagun, Mike Higgins, V~Ugur Guney, Volkan Cirik, and
  Kyunghyun Cho. 2017.
\newblock Searchqa: A new q\&a dataset augmented with context from a search
  engine.
\newblock \emph{arXiv preprint arXiv:1704.05179}.

\bibitem[{Fader et~al.(2014)Fader, Zettlemoyer, and Etzioni}]{openqa}
Anthony Fader, Luke Zettlemoyer, and Oren Etzioni. 2014.
\newblock Open question answering over curated and extracted knowledge bases.
\newblock In \emph{Proceedings of the 20th ACM SIGKDD international conference
  on Knowledge discovery and data mining}, pages 1156--1165. ACM.

\bibitem[{Joshi et~al.(2017)Joshi, Choi, Weld, and Zettlemoyer}]{triviaqa}
Mandar Joshi, Eunsol Choi, Daniel Weld, and Luke Zettlemoyer. 2017.
\newblock Triviaqa: A large scale distantly supervised challenge dataset for
  reading comprehension.
\newblock In \emph{Proceedings of the 55th Annual Meeting of the Association
  for Computational Linguistics (Volume 1: Long Papers)}, volume~1, pages
  1601--1611.

\bibitem[{Kratzwald and Feuerriegel(2018)}]{adaptive}
Bernhard Kratzwald and Stefan Feuerriegel. 2018.
\newblock Adaptive document retrieval for deep question answering.
\newblock \emph{arXiv preprint arXiv:1808.06528}.

\bibitem[{Kwiatkowski et~al.(2013)Kwiatkowski, Choi, Artzi, and
  Zettlemoyer}]{scaling}
Tom Kwiatkowski, Eunsol Choi, Yoav Artzi, and Luke Zettlemoyer. 2013.
\newblock Scaling semantic parsers with on-the-fly ontology matching.
\newblock In \emph{Proceedings of the 2013 conference on empirical methods in
  natural language processing}, pages 1545--1556.

\bibitem[{Kwiatkowski et~al.(2019)Kwiatkowski, Palomaki, Rhinehart, Collins,
  Parikh, Alberti, Epstein, Polosukhin, Kelcey, Devlin
  et~al.}]{naturalquestions}
Tom Kwiatkowski, Jennimaria Palomaki, Olivia Rhinehart, Michael Collins, Ankur
  Parikh, Chris Alberti, Danielle Epstein, Illia Polosukhin, Matthew Kelcey,
  Jacob Devlin, et~al. 2019.
\newblock Natural questions: a benchmark for question answering research.
\newblock \emph{Transactions of the Association for Computational Linguistics}.

\bibitem[{Lee et~al.(2018)Lee, Yun, Kim, Ko, and Kang}]{recall}
Jinhyuk Lee, Seongjun Yun, Hyunjae Kim, Miyoung Ko, and Jaewoo Kang. 2018.
\newblock Ranking paragraphs for improving answer recall in open-domain
  question answering.
\newblock \emph{arXiv preprint arXiv:1810.00494}.

\bibitem[{Lee et~al.(2016)Lee, Salant, Kwiatkowski, Parikh, Das, and
  Berant}]{rasor}
Kenton Lee, Shimi Salant, Tom Kwiatkowski, Ankur Parikh, Dipanjan Das, and
  Jonathan Berant. 2016.
\newblock Learning recurrent span representations for extractive question
  answering.
\newblock \emph{arXiv preprint arXiv:1611.01436}.

\bibitem[{Liang et~al.(2013)Liang, Jordan, and Klein}]{compositional}
Percy Liang, Michael~I Jordan, and Dan Klein. 2013.
\newblock Learning dependency-based compositional semantics.
\newblock \emph{Computational Linguistics}, 39(2):389--446.

\bibitem[{Lin(2019)}]{hype}
Jimmy Lin. 2019.
\newblock The neural hype and comparisons against weak baselines.
\newblock In \emph{ACM SIGIR Forum}.

\bibitem[{Logeswaran and Lee(2018)}]{quickthoughts}
Lajanugen Logeswaran and Honglak Lee. 2018.
\newblock An efficient framework for learning sentence representations.
\newblock \emph{arXiv preprint arXiv:1803.02893}.

\bibitem[{Mikolov et~al.(2013)Mikolov, Chen, Corrado, and Dean}]{skipgram}
Tomas Mikolov, Kai Chen, Greg Corrado, and Jeffrey Dean. 2013.
\newblock Efficient estimation of word representations in vector space.
\newblock \emph{arXiv preprint arXiv:1301.3781}.

\bibitem[{Narasimhan et~al.(2016)Narasimhan, Yala, and Barzilay}]{ie_external}
Karthik Narasimhan, Adam Yala, and Regina Barzilay. 2016.
\newblock Improving information extraction by acquiring external evidence with
  reinforcement learning.
\newblock \emph{arXiv preprint arXiv:1603.07954}.

\bibitem[{Perone et~al.(2018)Perone, Silveira, and Paula}]{elmopool}
Christian~S Perone, Roberto Silveira, and Thomas~S Paula. 2018.
\newblock Evaluation of sentence embeddings in downstream and linguistic
  probing tasks.
\newblock \emph{arXiv preprint arXiv:1806.06259}.

\bibitem[{Peters et~al.(2018)Peters, Neumann, Iyyer, Gardner, Clark, Lee, and
  Zettlemoyer}]{elmo}
Matthew~E. Peters, Mark Neumann, Mohit Iyyer, Matt Gardner, Christopher Clark,
  Kenton Lee, and Luke Zettlemoyer. 2018.
\newblock Deep contextualized word representations.
\newblock In \emph{Proc. of NAACL}.

\bibitem[{Radford et~al.(2019)Radford, Wu, Child, Luan, Amodei, and
  Sutskever}]{gpt2}
Alec Radford, Jeffrey Wu, Rewon Child, David Luan, Dario Amodei, and Ilya
  Sutskever. 2019.
\newblock Language models are unsupervised multitask learners.
\newblock \emph{OpenAI Blog}.

\bibitem[{Rajpurkar et~al.(2016)Rajpurkar, Zhang, Lopyrev, and Liang}]{squad}
Pranav Rajpurkar, Jian Zhang, Konstantin Lopyrev, and Percy Liang. 2016.
\newblock Squad: 100,000+ questions for machine comprehension of text.
\newblock In \emph{Proceedings of the 2016 Conference on Empirical Methods in
  Natural Language Processing}, pages 2383--2392.

\bibitem[{Robertson et~al.(2009)Robertson, Zaragoza et~al.}]{bm25}
Stephen Robertson, Hugo Zaragoza, et~al. 2009.
\newblock The probabilistic relevance framework: Bm25 and beyond.
\newblock \emph{Foundations and Trends in Information Retrieval},
  3(4):333--389.

\bibitem[{Singh(2012)}]{qaretrieval}
Amit Singh. 2012.
\newblock Entity based q\&a retrieval.
\newblock In \emph{Proceedings of the 2012 Joint conference on empirical
  methods in natural language processing and computational natural language
  learning}, pages 1266--1277. Association for Computational Linguistics.

\bibitem[{Taylor(1953)}]{cloze}
Wilson~L Taylor. 1953.
\newblock “{C}loze procedure”: A new tool for measuring readability.
\newblock \emph{Journalism Bulletin}, 30(4):415--433.

\bibitem[{Voorhees(2001)}]{trec01}
Ellen~M Voorhees. 2001.
\newblock Overview of the trec 2001 question answering track.
\newblock In \emph{In Proceedings of the Tenth Text REtrieval Conference
  (TREC}. Citeseer.

\bibitem[{Wang et~al.(2018)Wang, Yu, Guo, Wang, Klinger, Zhang, Chang, Tesauro,
  Zhou, and Jiang}]{r3}
Shuohang Wang, Mo~Yu, Xiaoxiao Guo, Zhiguo Wang, Tim Klinger, Wei Zhang, Shiyu
  Chang, Gerry Tesauro, Bowen Zhou, and Jing Jiang. 2018.
\newblock R 3: Reinforced ranker-reader for open-domain question answering.
\newblock In \emph{Thirty-Second AAAI Conference on Artificial Intelligence}.

\bibitem[{Yang et~al.(2017)Yang, Fang, and Lin}]{anserini}
Peilin Yang, Hui Fang, and Jimmy Lin. 2017.
\newblock Anserini: Enabling the use of lucene for information retrieval
  research.
\newblock In \emph{Proceedings of the 40th International ACM SIGIR Conference
  on Research and Development in Information Retrieval}, pages 1253--1256. ACM.

\bibitem[{Yang et~al.(2019)Yang, Xie, Lin, Li, Tan, Xiong, Li, and
  Lin}]{bertserini}
Wei Yang, Yuqing Xie, Aileen Lin, Xingyu Li, Luchen Tan, Kun Xiong, Ming Li,
  and Jimmy Lin. 2019.
\newblock End-to-end open-domain question answering with bertserini.
\newblock \emph{arXiv preprint arXiv:1902.01718}.

\end{thebibliography}
\bibliographystyle{acl_natbib}

\end{document}